\documentclass[twoside,leqno,twocolumn]{article}

\usepackage{ltexpprt}

\usepackage[ruled,vlined,algo2e]{algorithm2e}
\usepackage{amsmath}
\usepackage{amssymb}
\usepackage{booktabs}
\usepackage[space]{cite}
\usepackage{graphicx}
\usepackage{hyperref}
\usepackage{marvosym}
\usepackage{multirow}
\usepackage{xcolor}
\usepackage{zi4}
\usepackage{bold-extra}
\usepackage{breqn}
\usepackage{placeins}

\definecolor{blue}{HTML}{2980b9}
\definecolor{gray}{HTML}{7f8c8d}
\definecolor{green}{HTML}{16a085}
\definecolor{orange}{HTML}{d35400}
\definecolor{purple}{HTML}{8e44ad}
\definecolor{red}{HTML}{c0392b}

\begin{document}

\title{\Large Efficient Wind Speed Nowcasting with GPU-Accelerated Nearest Neighbors Algorithm}
\author{
  Arnaud Pannatier  \thanks{Idiap Research Institute, \url{arnaud.pannatier@idiap.ch} \Letter } \Letter
  \and Ricardo Picatoste \thanks{SkySoft-ATM \url{picatoro@skysoft-atm.com}}
  \and François Fleuret  \thanks{Université de Genève \url{francois.fleuret@unige.ch}}
}
\date{}
\maketitle

\fancyfoot[R]{\scriptsize{Copyright \textcopyright\ 2022 by SIAM\\
    Unauthorized reproduction of this article is prohibited}}

\providecommand{\keywords}[1]
{
  \small
  \textbf{\textit{Keywords---}} #1
}

\begin{abstract}

  \small\baselineskip=9pt
  This paper proposes a simple yet efficient high-altitude wind nowcasting pipeline.
  It processes efficiently a vast amount of live data recorded by airplanes over the whole airspace and reconstructs the wind field with good accuracy.
  It creates a unique context for each point in the dataset and then extrapolates from it. As creating such context is computationally intensive,
  this paper proposes a novel algorithm that reduces the time and memory cost by efficiently fetching nearest neighbors in a data set whose elements are organized along smooth trajectories that can be approximated with piece-wise linear structures.
  We introduce an efficient and exact strategy implemented through algebraic tensorial operations, which is well-suited to modern GPU-based computing infrastructure.
  This method employs a scalable Euclidean metric and allows masking data points along one dimension.
  When applied, this method is more efficient than plain Euclidean $k$-NN and other well-known data selection methods such as KDTrees and provides a several-fold speedup.
  We provide a PyTorch implementation and a novel data set to replicate empirical results.
\end{abstract}


\normalsize
\section{Introduction}
High-Altitude Wind speed nowcasting is crucial for air traffic management, as it directly impacts the behavior of airplanes.
Even if the accuracy of numerical weather prediction still increases at an incredible pace \cite{Bauer2015}, and that traditional methods are well-suited for mid-long term forecasting (between 6h and 14 days), it remains less accurate than extrapolating the last measurements in the first few hours \cite{Pulkkinen2019}.
Major actors are developing new weather nowcasting pipelines \cite{DeepMind2021} where the overall strategy for all nowcasting persists in getting good quality measurements and extrapolate.
These measurements are available in the case of high-altitude wind speed nowcasting, as airplanes record the wind speed along their trajectory.
To predict the wind at a given time and space, one needs to extrapolate from contexts made of the last measurements recorded in the vicinity of the desired point.
Simple machine learning methods such as $k$-Nearest Neighbors ($k$-NN) can be employed to build such contexts.
However, the required linear search for such approaches is demanding both in terms of computation and memory footprint, to the point that it may make them prohibitively expensive.
A standard strategy used to reduce this cost consists of partitioning the space while relying on exact or approximate criteria to reject groups of samples,
cutting the temporal and memory cost to a sub-linear function of their number.
This is used for example in Locality Sensitive Hashing (LSH) \cite{Gionis1999}, or other tree-based methods \cite{Bentley1975,Yianilos1970}.

Besides the computational aspect, an essential characteristic of $k$-NN is the soundness of the distance function.
Considering both the units and dynamic ranges of distinct features differ, their combination may be meaningless.
A sound and straightforward strategy to address this issue is to scale features individually.
The scaling factors can come from prior knowledge (\emph{e.g.} sampling rates, instrument calibration, physical laws) or can be optimized from the data to maximize the performance.
When training a model with temporal data, another problem arises: to make a forecast at any time, we have to exclude data points in the future, or according to a similar criterion, that may, for instance, integrate additional constraints related to processing time.
This implies that some processing must be done before using $k$-NN and possibly for each data set point.

With this in mind, we propose a novel algorithm called Trajectory Nearest Neighbors (TNN) that tackles these two problems while being particularly adapted to point data sets that can be covered with cylinders, such as smooth trajectories.
Each of these cylinders can be represented by a segment and an error distance term.
In priority, our method explores points that belong to the nearby cylinders. It stops when all the nearest neighbors are guaranteed to be retrieved. Our approach is exact and uses basic linear algebra subprograms (BLAS) \cite{cuBLAS} that can be computed efficiently on GPU using standard frameworks \cite{Paszke2019, Martin2016}.
This algorithm's strength comes from its simplicity while still offering a substantial increase in performance when used in the correct setup.

This paper also presents a high-altitude wind nowcasting pipeline while comparing it with baselines and a particle model developed to reconstruct wind fields using a similar data set \cite{Sun2017}.
Using TNN, we trained our model over a data set of 35 days and over 33 million points.
Furthermore, our model managed to beat the other methods evaluated.
We analyzed the speed up of our approach, and it shows that in the context of GPU computing, our method seems to give the best speed up compared to linear search and traditional methods while avoiding using any other libraries or language.
We also demonstrate that when a data set cannot be well separated into trajectories, our approach does not offer a substantial increase in performance.
With this approach, we could speed up our pipeline's training by two orders of magnitude, making it achievable in a reasonable time. The main contributions of this paper are the following:
\begin{itemize}
  \item The introduction of a novel algorithm called Trajectory Nearest Neighbors (TNN) based on simple linear algebra and its theoretical analysis.
  \item An extensive comparison with traditional approaches (linear search,
        KDTrees.)
  \item A concrete application in the context of high-altitude wind nowcasting.
  \item An implementation of this algorithm using PyTorch \cite{Paszke2019}.
  \item The publication of a data set containing 33 million points measured along plane trajectories over the course of 35 days. \footnote{Code and datasets are available at \href{https://github.com/idiap/tnn}{github.com/idiap/tnn}}
\end{itemize}

\section{Related Works}
Our works differ from other nowcasting pipelines that often remain oriented towards the forecasting of radar products such as rainfall \cite{DeepMind2021, Shi2017, NowDeepN2021} as it can leverage the grid structure of the data and use standard computer vision strategies.
Other works were done on similar datasets \cite{Sun2017, Kikuchi2018}, we compared our models to the first one that reconstructs the wind field by using a particle model. The second one uses these measurements to select the best subset of an ensemble of weather forecasts.
There were recent works designed to fetch nearest neighbors using CUDA \cite{cuda} \cite{Garcia2010, SweetKNN2017}. However, they require a Euclidean metric and no temporal coordinates, making them hard to compare to our case.

\section{Methods and Technical Solutions}
\subsection{General Description of Wind Nowcasting Pipeline}

\begin{figure}
  \centering
  \includegraphics[scale=0.20]{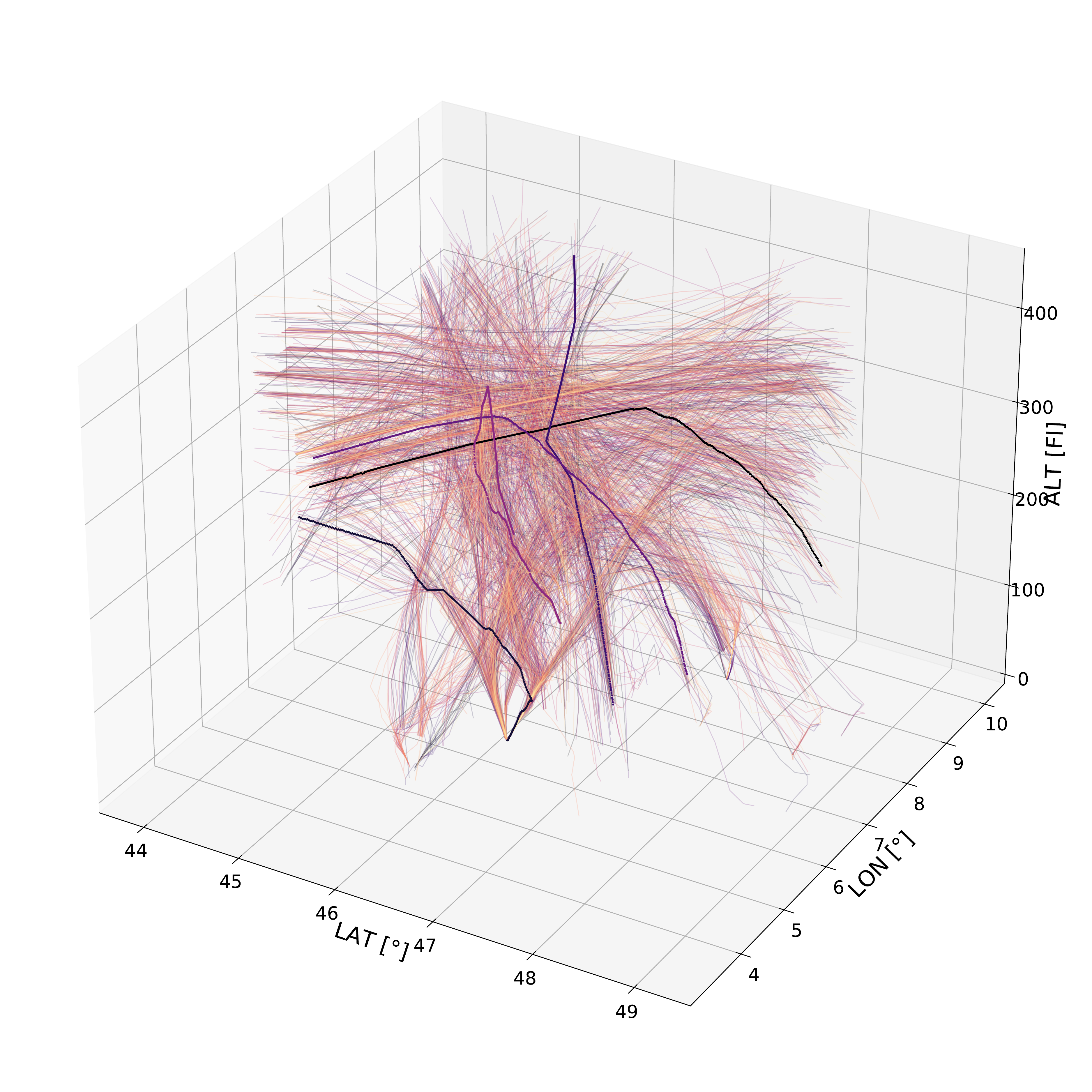}
  \caption{ All the plane trajectories recorded on Day \#1 are plotted.
    Most of them are clustered along some main routes.
    A few trajectories are highlighted, which show the density of the tracking points.}
  \label{fig:dataset}
\end{figure}

We have at our disposal Mode-S data recordings for 35 different days over European airspace [Fig.\ref{fig:dataset}].
Mode-S data is exchanged between Secondary Surveillance Radars (SSR) and the aircraft radar transponders \cite{enwiki:modes},
and contains, among others, the position of the plane and measurements of the wind.
SSRs rotate with a period of 4 seconds, setting the sampling time for these variables.
The pipeline for training the model is the following: batches of random samples are sampled iteratively. For each of these samples, a context of past measurements is retrieved, where each context contains only points that were measured at least $t_{w}$ minutes in the past ($t_{w}$ is typically 30 minutes) and from which the forecasts are then extrapolated. We use \textit{Root Mean Square Error (RMSE)} to measure models' performance.

\subsection{High-Altitude Wind Nowcasting}

We extrapolate the wind speed measurements using Gaussian Kernel Averaging (GKA) \cite{friedman2001} to give the forecast $\hat{s}$ at point $(x,y,z,t)$:

\begin{equation} \label{eq:gka}
  \hat{\vec{s}}(x, y, z, t) =   \frac{\sum\limits_{k \in \mathcal{C}} e^{
        \sigma_{xy} \left[ \left( x\!-\!x_{k} \right)^2 + \left( y\!-\!y_{k} \right)^2 \right]
        +\sigma_{z} \left( z\!-\!z_{k} \right)^2
        +\sigma_t \left( t\!-\!t_{k} \right)^2
      } \vec{ s_{k}}}{\sum\limits_{k \in \mathcal{C}} e^{
        \sigma_{xy} \left[ \left( x\!-\!x_{k} \right)^2 + \left( y\!-\!y_{k} \right)^2 \right]
        +\sigma_{z} \left( z\!-\!z_{k} \right)^2
        +\sigma_t \left( t\!-\!t_{k} \right)^2
      }}
\end{equation}
Where $\vec{s_k}$ is the wind speed measured at $(x_k,y_k, z_k, t_k)$, $(\sigma_{xy}, \sigma_{z}, \sigma_{t})$ are parameters of the model and $\mathcal{C}$ is the context of the point.

The parameters $\vec{\sigma}$  can be physical constants found either through optimization or through careful geostatistical analysis.
$\vec{\sigma}$ can also be modelled as a function of the space $\vec{\sigma} \colon (x,y,z) \mapsto (\sigma_{xy}, \sigma_z, \sigma_t)$, using a Multi-Layer Perceptron (MLP) which enables the model to resize the weights anywhere on the space.
For example, in a region with numerous measurements, it could be wise to reduce the scaling factor to put more weight on only a few closer neighbors.
Conversely, increasing the scaling might be better to average more points in a region where the measurements are sparse.

The context $\mathcal{C}$ can contain \textit{a priori} all the previous measurements, which makes the computation prohibitively expensive.
To make the training feasible, we decided to reduce the number of points by only taking the few contributing the most to the sum.
This approach is reasonable as most points will only have a negligible contribution to the final weighted average.

\subsection{Partitionning Measurements in Segments}

\begin{figure}
  \centering
  \includegraphics[scale=1]{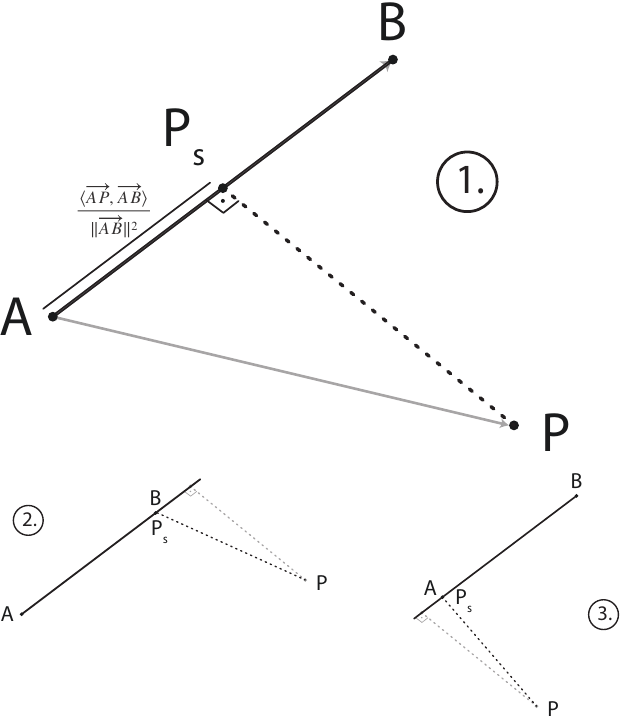}
  \caption{
    Illustration of the projection and the clamping process used to find the distance between one point $P$ and a segment $s$. 1. If the projection of $P$ falls between the point starting point $A$ and endpoint $B$, no clamping is needed, and we can retrieve the points using a simple geometric formula. 2-3. If the projection falls outside the segment, the closest point in $s$ would be $A$ or $B$.
  }
  \label{fig:geometry}
\end{figure}

Retrieving a context of points involves finding their nearest neighbors in the dataset recorded before a specific time.
The traditional nearest neighbors algorithm typically partitions the space and explores only a tiny number of partitions.
Given that our measurements are clustered along some routes [Fig. \ref{fig:dataset}] and that we have to filter if according to a temporal criterion, the traditional approaches such as Locality Sensitive Hashing (LSH) \cite{Gionis1999}, KDTrees \cite{Bentley1975}, Ball Trees \cite{Omohundro1989}, and Vantage Trees \cite{Yianilos1970} are not well suited. Moreover, because masking is complex to implement for these structures, the algorithm explores many partitions containing only non-valid measurements.
When working with data recorded along trajectories, the natural way of splitting the space is to split these trajectories into segments. Masking can then be done efficiently by only looking at the timestep of the first point associated with the segment.
Once the measurements are split into segments, one can filter the valid segments and retrieve the nearest neighbors by only exploring nearby segments.

\subsection{Theoretical Analysis} \label{sec:theory}
We have a data set of $N$ measurements from wind speed at location $\vec{x_i} = (x_i,y_i,z_i, t_i) \in \mathbb{R}^4, i \in \{1, \dots N \}$ and planes only record the $x,y$ components of the wind speed $\vec{s}_i = (s_i^{x}, s_i^{y}) \in \mathbb{R}^2, i \in \{ 1, \dots N \}$.
For the sake of simplicity, we split the measurements location $\vec{x}_i$ into a point $P_i = (x_i,y_i,z_i) \in \mathbb{R}^3$ and a time-step $t_i$.
We will use the following scalable metric :

\begin{dmath}
  \lVert \vec{x_1}-\vec{x_2} \rVert_{\vec{\sigma}}^2  =
  \sigma_{xy} [ (x_1-x_2)^2 + (y_1-y_2)^2 ] + \sigma_z ( z_1 - z_2 )^2 + \sigma_t (t_1-t_2)^2
\end{dmath}
This is split into a spatial and a temporal part:
\begin{align}
  \lVert P_1-P_2 \rVert^2_{\sigma_{xyz}} & =  \sigma_{xy} [ (x_1-x_2)^2 + (y_1-y_2)^2 ] + \sigma_z ( z_1 - z_2 )^2 \\
  \rVert t_1-t_2 \rVert ^2_{\sigma_{t}}  & = \sigma_t (t_1-t_2)^2
\end{align}

Where $\vec{\sigma} = \left( \sigma_{xy},\sigma_z,\sigma_t \right)$ is a scaling factor that allows a correct comparison between dimensions.
All the points in the data set belong to trajectories that we split into sets $T_j = \{\vec{x}_{j,1},...,\vec{x}_{j,K} \}$, $ j \in \{ 1, \dots, \frac{N}{K} \}$ of exactly $K$ elements and where the points are sorted along the time dimension. We pad the last set with copies of the first and last elements if the cardinality of the sets is not $K$. Each set $T_j$ can be approximated by a segment $s_j$ between the first point ($A$) and the last point ($B$) of the set,
where $A$ and $B$ correspond to the spatial part of the measure $\vec{x}_{j,1}$ and $\vec{x}_{j,K}$.

The minimum distance $d$ between a point $P$ recorded at time $t$ and a segment $s$ is given by:
\begin{equation}
  d = \text{dist}((P,t), s) = \lVert P-P_s \rVert^2_{\sigma_{xyz}} + \lVert t-t_s \rVert ^2_{\sigma_{t}}
\end{equation}

where $P_s$ is the projection of $P$ on the segment $AB$, and $t_s^w$ is the projection of $t$ on $[t_A,t_B]$ [Fig. \ref{fig:geometry}]:
\begin{align} \label{eq:projection-segment}
  P_s   & = A + \max \left( 0, \min{\left(  \frac{\langle \overrightarrow{AP} ,\overrightarrow{AB}  \rangle}{ \lVert \overrightarrow{AB}  \rVert^2_2},1 \right)} \right) \overrightarrow{AB} \\
  t_{s} & = \max(t_A, \min(t,t_{B})) \label{eq:temporal-part}
\end{align}

When approximating a subtrajectory $T = \{ \vec{x}_{1},...,\vec{x}_{K} \}$ by a segment $s$, we are making an approximation error for each point $i \in \{ 1, \dots, K \}$ which can be upper bounded :
\begin{align}
  E_{\text{app}} & = \max_{i \in 1,\dots,K}{ \text{dist}(\vec{x_{i}},s)}                                                                     \\
                 & = \max_i{\lVert P_i-P_{s_i} \rVert_{\sigma_{xyz}}^2 + \underbrace{\lVert t_i-t_{s_i} \rVert^2_{\sigma_t}}_{0} } \nonumber \\
                 & \leq \sigma_{max}  \max_i \lVert P_i-P_{s_i} \rVert_2^2 \nonumber
\end{align}
with $\sigma_{max} = \max(\sigma_{xy}, \sigma_{z})$ and where the temporal part of the distance vanishes as $t_{A} < t_{i} < t_{B}$ by construction. This upper bound can be computed efficiently, as only the maximum Euclidean distance of the points to the segments is required.

By slightly modifiying equation \ref{eq:temporal-part}, one can mask all the segments that contains points only measured after a certain time window $t_{w}$ (typically 30 minutes) before $t$  :
\begin{equation}
  t_s^w  = \left\{ \begin{matrix} \infty      & \text{if } t_A > t-t_w \\  \label{eq:temporal-part-masked}
               \min(t,t_B) & \text{otherwise}
  \end{matrix} \right.
\end{equation}

The masked distance is then given by:
\begin{equation}
  d_{w} = \lVert P-P_s \rVert^2_{\sigma_{xyz}} + \lVert t-t_s^w \rVert ^2_{\sigma_{t}} \label{eq:distance}
\end{equation}

If we take into account the error that we are making when approximating a set of points by a segment $s$, we can know that all these points are at least at a distance $(d_{w}-E_{\text{app}})$ from a point $P$. Thus, this criterion can be used to exclude segments in search of neighbors if one already has a candidate for the $k$-th neighbors that are closer than $d_{w} - E_{\text{app}}$.

\begin{figure}
  \centering
  \includegraphics[scale=1.6]{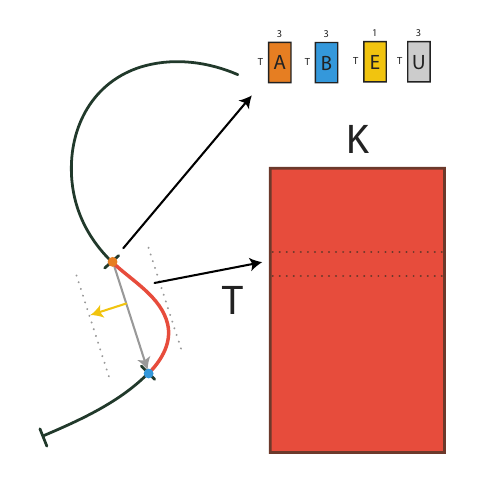}
  \caption{Each trajectory is split into sub-trajectories (here in red) that are approximated by a segment (gray) and an error term (yellow). A $T \times K$ matrice holds the indices of all points belonging to each segments. To be able to compute the distance efficiently, we group the segments' information in matrices $\mathbf{A},\mathbf{B}, \mathbf{E}, \mathbf{U}$.}
  \label{fig:2dtrajectorycreator}
\end{figure}

\subsection{Trajectory Nearest Neighbors Algorithm (TNN)} \label{sec:algorithm}

The algorithm takes as an input a batch of $M$ points and for each retrieves its $k$ nearest neighbors. It is written in a tensorial way, making extensive use of broadcasting, which is efficiently implemented in standard libraries such as PyTorch \cite{Paszke2019} and TensorFlow \cite{Martin2016}.
The first task, described in Algorithm \ref{alg:compute-distance}, is to compute the distance from all the batch points to all the segments, using the formula [Eq. \ref{eq:distance}] while taking the approximation error into account. If the approximation error is larger than the estimated distance from the point to the segment, this distance is set to zero.
The information required to compute the distance to a segment are its starting point $A$ measured at time $t_A$, ending point $B$ measured at time $t_b$ and the error term $E$. As the unitary vector $\vec{u} = \frac{\overrightarrow{AB}}{ \lVert \overrightarrow{AB}  \rVert^2_2}$ needs to be known for all queries, one can precompute as well and store its value [Eq. \ref{eq:projection-segment}]. All these quantities are arranged in matrices $\mathbf{A},\mathbf{t_A},\mathbf{B},\mathbf{t_B}, \mathbf{U}, \mathbf{E}$ to allow efficient broadcasting [Fig. \ref{fig:2dtrajectorycreator}]. A pseudocode version showing how this formulas can be broadcasted can be found in the supplementary, or in the actual version of the code.
\begin{algorithm2e}[!ht]
  \SetAlgoLined
  \KwData{$P$, $t$,$\vec{\sigma}$, $A$, $t_A$, $B$,$t_B$,$U$, $E$, $t_w$}
  \KwResult{Distance from $P$ to segment with error}
  \lIf{$t_A > t-t_w$}{
    \KwRet{ $\infty$}
  }
  $\delta = \text{clamp}((P-A) \cdot U,0,1)$\;
  $P_{s} = A + \delta * (B-A)  $ \;
  $D = \lVert P_{s} - P \rVert ^2_{\sigma_{xyz}} + \sigma_t\text{max}(t_B-t, 0)^2$  \;
  $D =  D - E\max(\sigma_{xy},\sigma_{z})$\;
  \KwRet{ clamp$(D, 0)$}
  \caption{Distance from point $P$ to a segment, details about the notation can be found in section \ref{sec:algorithm}}
  \label{alg:compute-distance}
\end{algorithm2e}
\begin{algorithm2e}[!ht]
  \SetAlgoLined
  \KwData{A batch of point, informations about the segments}
  \KwResult{$k$-Nearest Neighbors}
  Distances = compute distances to segments (\ref{alg:compute-distance})\;
  Distances $=$ sort distances\;
  Furthest neighbors $= \infty$\;
  Next distance $=$  Distances$[:,0]$\;
  i = 1\;
  d = 0\;
  \While{Furthest neighbors $\geq$ Next distance or $d=M$}{
    Fetch $F$ segments of $K$ points for the remaining ($M-d$) points in the batch\;
    Compute distance from  batch points to segments points\;
    Current nearest neighbors $=$ sort previous ($k$) and new points ($FK$)\;
    Furthest neighbor = Current nearest neighbors[:, k]\;
    $d$ = nb of completed lines\;
    Put completed lines ($d$) at the end of the batch \;
    Next distances $=$ Distances$[:,i*F]$\;
    i += 1\;
  }
  \KwRet{$k$-Nearest Neighbors}
  \caption{Trajectory Nearest Neighbors}
  \label{alg:main-loop}
\end{algorithm2e}

Once the first task is completed, the algorithm searches for the neighbors in the closest $F$ segments [Alg. \ref{alg:main-loop}], which consists of fetching all the points contained in a few of them at a time, then computing the actual point-to-point distance between the points in the batch and all the points contained in the segments.
At that stage, points that are recorded after a time windows $t_{w}$ before $t$ are masked. The algorithm then sorts the point by distance and keeps only the first $k$ ones.
At this point, it has $k$ candidates for the nearest neighbors and has to decide if it is necessary to continue the search.
If the distance from the furthest candidate is smaller than the distance to the next segment while taking the error term into account, it is guaranteed that no other points can be closer, which means that the neighbors are found.
If the algorithm is already completed for some points in the batch, it puts them at the end of the batch and continues to process only the remaining points.
For that, it fetches points in the next closest segments until the following segments are too far to include any valid candidates.
This continues until the neighbors for all the points in the batch are found.

\subsection{Performance and Memory analysis} \label{sec:perf-mem}

By splitting the data set into segments, TNN reduces the number of comparisons needed to find the nearest neighbors.
The following table describes the number of operations and the memory required by the linear search and the TNN algorithm [Tab. \ref{tab:complexity-storage-details}].
To compute the linear search, we used the well-known distance matrix tricks \cite{Albanie2019}.
In both cases, the bottleneck is the top-$k$ algorithm and the sorting subroutine needed to find the nearest neighbors.
The distance to segment subroutine [Alg. \ref{alg:compute-distance}] requires computing the distance matrix from all points to the segments, using [Eq.\ref{eq:distance}].
As the number of segments is roughly $\frac{N}{K}$, this offers a substantial improvement over the linear search.
Let us define the mean number of segments $n_f$ that we have to consider for one point in the nearest neighbors' search. Hence the algorithm has to fetch a $n_f \times F \times K$ points for all $N$ points.
This factor depends heavily on the data set. For example, it is lower when the segments are a good approximation of the trajectories and when they are well separated.
In the following section, we illustrate how this factor's variability applies in an experiment with synthetic data through consideration of both a usual and a pathological case
One can see that our method offers a substantial decrease in storage footprint as well.
This is crucial when working with GPU as the memory is often limited.
We computed the theoretical cost for a single batch as most of the memory can be freed directly after usage.
\begin{table*}[ht]
  \caption{
    Time-complexity and Space-Complexity estimation for computing the nearest neighbors of all points in the data set.
    $N$ is the size of the data set. $n_f$ is the mean number of times we have to fetch all $K$ points in $F$ segments in the nearest neighbors' search.
    The storage is described for a batch of $M$ points as most space can be directly freed after the operation.
    The total considers only the sorted matrix and the top-k neighbors as it is what the memory contains at peak use.
  }
  \label{tab:complexity-storage-details}
  \centering
  \begin{tabular}{llll}
    \toprule
    \bf{Method}                    & \bf{Steps}         & \bf{Time Complexity}                         & \bf{Space Complexity}      \\\midrule
    \multirow{2}{*}{Linear search} & Distance Matrix    & $O(N^2D)$                                    & $M(2N+1)$                  \\
                                   & Top-k              & $O(N^2)$                                     & $2Mk$                      \\\midrule
    \multirow{5}{*}{TNN}           & Distance Segments  & $O(\frac{N^2}{K}D)$                          & $M\frac{N}{K}D$            \\
                                   & Sort               & $O(\frac{N^2}{K} \log(\frac{N}{K}))$         & $2M \frac{N}{K}$           \\ \cmidrule{2-4}
                                   & Distance to points & $O(Nn_fFKD)$                                 & $M(k+FK)D $                \\
                                   & Top-k              & $O(Nn_f(k+FK))$                              & $2M(k+FK)$                 \\
                                   & Total              & $O(\frac{N^2}{K} \log(\frac{N}{K})+Nn_fFKD)$ & $M\frac{N}{K}D + M(k+FK)D$ \\
    \bottomrule
  \end{tabular}
\end{table*}

\section{Empirical evaluation}

The wind-speed nowcasting pipeline and the trajectory nearest neighbors algorithm (TNN) are evaluated in different settings. Some baselines are used to compare the performance of our model, while TNN is tested against some other nearest neighbors algorithms.

\subsection{High-Altitude Wind Nowcasting}

\begin{table*}
  \caption{
    Results of the different models for three days in the data set, where the mean wind speed is specified for each day.
    The baselines are in \textit{italic} and the total duration for one epoch is mentioned for a single day dataset (around 1mio measurements) and for a five-week dataset (around 33mio measurements).
    The $k$-NN and GKA baselines' optimizations are done with SciKit-Opt using Random Forest method.
    Our methods are optimized with Adam using the whole training set, but the averaging set for each point is restricted by TNN, which reduces the training time by about two orders of magnitude.
  }
  \label{tab:results}
  \centering
  \begin{tabular}{lrrrrr} \toprule
                                  & \multicolumn{3}{c}{\textbf{RMSE [kn]} } & \multicolumn{2}{c}{\textbf{Epoch duration } }                                                                        \\
    \textbf{Model}                & \textbf{Day \#1}                        & \textbf{Day \#2}                              & \textbf{Day \#3} & \textbf{1 day dataset} & \textbf{5 weeks dataset} \\
    Mean wind                     & 95 [kn]                                 & 49 [kn]                                       & 39 [kn]          & \tt{hh:mm:ss}          & \tt{hh:mm:ss}            \\ \midrule
    \textit{Day Average}          & 27,87                                   & 20,19                                         & 13,86            & \tt{0:03}              & \tt{2:05}                \\
    \textit{Hour Average}         & 26,19                                   & 17,51                                         & 12,67            & \tt{0:34}              & \tt{20:00}               \\
    Particle Model \cite{Sun2017} & 9,98                                    & 10,07                                         & 7,84             & \tt{6:57:15}           & \texttt{1121:54:30}      \\
    \textit{GKA}                  & 9,07                                    & 9,64                                          & 7,66             & \tt{2:39:18}           & \tt{481:47:20}           \\
    \textit{$k$-NN | Persistence} & 9,02                                    & 9,86                                          & 7,57             & \tt{4:31:47}           & \tt{558:37:05}           \\
    GKA - TNN                     & 8,71                                    & 9,19                                          & 7,55             & \tt{\textbf{4:13}}     & \tt{\textbf{1:35:30}}    \\
    \bf{GKA - MLP - TNN}          & \bf{8,01}                               & \bf{8,51}                                     & \bf{6,87}        & \tt{\textbf{4:21}}     & \tt{\textbf{1:37:39}}    \\
    \bottomrule
  \end{tabular}

\end{table*}

In table \ref{tab:results}, some simple baselines, a particle model \cite{Sun2017} and our GKA models are evaluated.
We use Root Mean Square Error (\textit{RMSE}) to assess the accuracy of our forecasts.
We start with simple and understandable models as baselines: The average wind of the whole day and the last hour before the prediction point.
Then we evaluate $k$-NN, where $k$ is optimized to give the best accuracy on a validation set. When $k = 1$, this model is equivalent to the persistence model.
For the day average model, we used a non-causal approach and predicted for each point in the test set the mean wind speed value for the day.
The first causal baseline we considered is the hour
average model. This model outputs the mean of all the points measured in the interval $[t-\text{1h30};t-\text{30min}]$ for the prediction of a given point a time $t$.
As expected, those baselines are far from optimal, but they give an informative way of comparing our model.
Training the naive $k$-NN and GKA baselines was feasible on a single data day, but it becomes intractable to train on the whole dataset without using a local context.
The particle model \cite{Sun2017} is giving results similar to our baseline, but
it probably could be improved by optimizing the hyperparameters of this model.
TNN allowed us to retrieve a small number of neighbors based on a scaled and masked metric.
This procedure made the training over the whole data set possible while beating the existing baselines' performance.
The time taken to evaluate all the points in the dataset is also mentioned in [Tab. \ref{tab:results}]. It shows that using TNN to build a proper context is mandatory to train our models by offering a speedup of 320 times compared to the fastest baseline.
Training models on the whole dataset is mandatory to test more advanced strategies. And when letting the parameters of GKA vary through space, we managed to increase the precision of our model.

\subsection{Comparison with Linear Search}
\begin{figure}
  \centering
  \includegraphics[scale=.275]{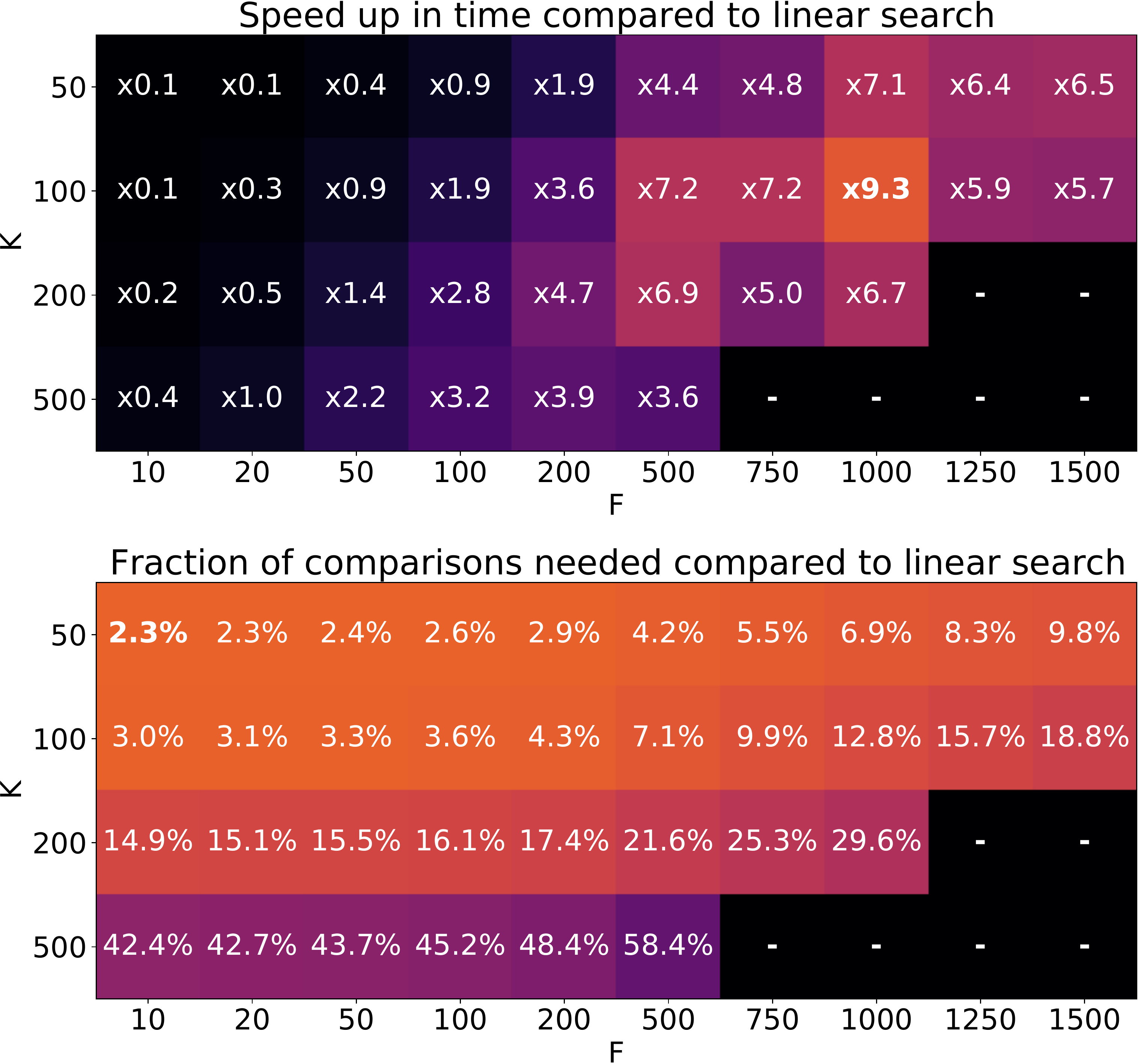}
  \caption{
    A comparison between linear search and TNN is made on GPU for different parameters $K$ (number of points per segment) and $F$ (number of segments fetched at the same time) as explained in section \ref{sec:algorithm}.
    The first grid refers to the average speedup reached by TNN compared to linear search (higher is better).
    The second grid shows the average percentage of comparisons that TNN needed to find the nearest neighbors (lower is better).
  }
  \label{fig:wind1000cuda}
\end{figure}

A comparison between TNN and Linear Search is made on two synthetic data sets and a real one.
The first one consists of smooth random walks. The second represents a pathological case where all points are taken randomly and are considered to be measured simultaneously. The final one uses actual wind speed measurements.
Our method substantially increases the performance by drastically restricting the search space [Fig. \ref{fig:wind1000cuda}][Tab. \ref{tab:numberofcomparison}].
The average number of comparisons on CPU is approximately thirty times lower than the linear search.
Changing the number of segments $F$ that are fetched simultaneously in Algorithm \ref{alg:main-loop} impacts the speed up time,
as when bringing many segments at a time, more comparisons can be made simultaneously, but it may perform some unnecessary ones, explaining why there is a sweet spot for the average speed up in figure \ref{fig:wind1000cuda}.
The same sweet spot effect can be observed by increasing the number of measurements per segment $K$.

Our method substantially increases the efficiency on CPU by dividing the search time by almost one order of magnitude.
We see that running the linear search on GPU instead of CPU already offers a substantial decrease in query time due to the massive parallelization properties of GPUs.
TNN exploits this property well, making it run sixteen times faster than linear search on GPU.
We tested the algorithm on the smoothed random walk and the original data set [Tab. \ref{tab:numberofcomparison}].
The trajectories are well separated in both data sets, and TNN increases the efficiency similarly.
We evaluate a pathological case where points are randomly distributed over the space and randomly grouped in segments, leading to a significant approximation error $E_{\text{app}}$. In that setting, while querying for nearest neighbors, the algorithm cannot filter far away segments and thus, resorts to checking all the points. In that case, TNN uses the same number of comparisons as the linear search and does not offer any speed up as expected. Indeed, the approximation error is significant in such a case, as the algorithm will have to consider almost all segments in the data set.
One solution to this problem could be reorganizing the data points so segments can approximate the resulting sets more appropriately.

\begin{table*}
  \caption{Mean number of comparisons and time needed for querying 1000 nearest neighbors on the original wind and smoothed random walk (SRW) data sets. As TNN is an exact method, the query results are the same for both methods. It shows as well the total duration (\texttt{hh:mm:ss}) needed to retrieve the neighbors for the whole data set. All these comparisons are made on a data set of a million points. }
  \label{tab:numberofcomparison}
  \centering
  \begin{tabular}{cclrrr}
    \toprule
    \bf{Data set}                      & \bf{Device}               & \bf{Algorithm} & \bf{Comparisons}          & \bf{Query [ms]}        & \bf{Total duration}     \\
    \midrule

    \multirow{4}{*}{Original Data set} & \multirow{2}{*}{\tt{CPU}} & Lin. Search    & \tt{811'372}              & \tt{9.03}              & \tt{2:30:32}            \\
                                       &                           & TNN            & \texttt{\textbf{28'579}}  & \tt{1.03}              & \tt{17:08}              \\
                                       & \multirow{2}{*}{\tt{GPU}} & Lin. Search    & \tt{811'372}              & \tt{2.55}              & \tt{42:28}              \\
                                       &                           & TNN            & \tt{81'611}               & \texttt{\textbf{0.16}} & \texttt{\textbf{2:43}}  \\\midrule
    \multirow{4}{*}{SRW Data set}      & \multirow{2}{*}{\tt{CPU}} & Lin. Search    & \tt{1'000'000}            & \tt{11.95}             & \tt{3:19:09}            \\
                                       &                           & TNN            & \texttt{\textbf{58'408}}  & \tt{1.60}              & \tt{26:35}              \\
                                       & \multirow{2}{*}{\tt{GPU}} & Lin. Search    & \tt{1'000'000}            & \tt{2.51}              & \tt{41:48}              \\
                                       &                           & TNN            & \tt{93'555}               & \texttt{\textbf{0.39}} & \texttt{\textbf{6:28}}  \\\midrule
    \multirow{4}{*}{Random points}     & \multirow{2}{*}{\tt{CPU}} & Linear Search  & \tt{1'000'000}            & \tt{7.43}              & \tt{2:03:51}            \\
                                       &                           & TNN            & \tt{999'940}              & \tt{15.51}             & \tt{4:18:34}            \\
                                       & \multirow{2}{*}{\tt{GPU}} & Linear Search  & \tt{1'000'000}            & \tt{1.72}              & \tt{28:42}              \\
                                       &                           & TNN            & \texttt{\textbf{998'588}} & \texttt{\textbf{1.36}} & \texttt{\textbf{22:40}} \\
    \bottomrule
  \end{tabular}
\end{table*}

\begin{figure}
  \centering
  \includegraphics[scale=1]{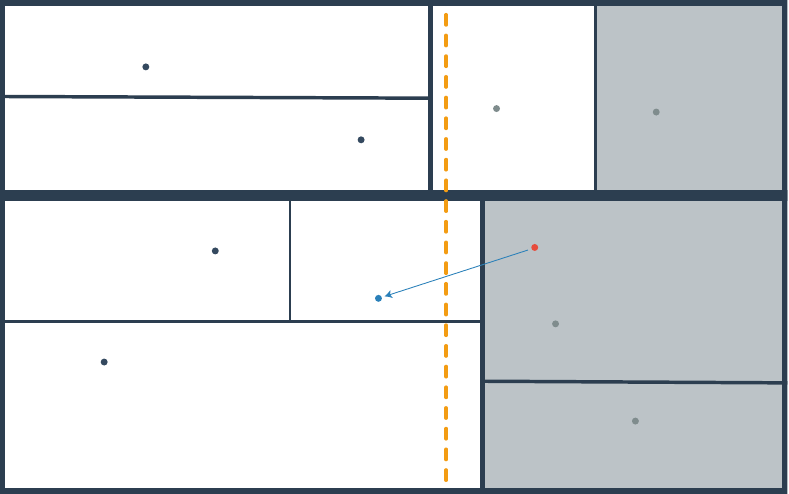}
  \caption{ The KDTree algorithm explores in priority cells that are close to the query point.
    Setting a hard limit on a given axis complicates the procedure as the cells close to the goal might not contain any valid candidate.
    The black lines represent the KDTree structure.
    The query point is depicted in red, and the nearest neighbor that respects the mask (in orange) is blue.
    The cells that should not be considered are colored in grey.
  }
  \label{fig:kdtree_masked}
\end{figure}

\begin{table*}[t]
  \caption{Comparison with KDTrees and linear search on GPU and CPU for the original data set. The creation of the different structures has to be performed once at the beginning of the program.
    The comparison details are given in section \ref{sec:kdtree}. It shows the average time for querying 1000 nearest neighbors on the different data sets and the total duration (\texttt{hh:mm:ss}) needed to retrieve the neighbors for the whole data set. All these comparisons are made on a data set of a million points.
  }
  \label{tab:comparison}
  \centering
  \begin{tabular}{lrrr}
    \toprule
    \bf{Method}           & \bf{Creation [s]} & \bf{Query [ms]}        & \bf{Total duration}     \\\midrule
    Linear search CPU     & \tt{- }           & \tt{9.03}              & \tt{2:30:32}            \\
    TNN CPU               & \tt{1.00}         & \tt{1.03}              & \tt{17:08}              \\
    Scaled masked KDTree  & \tt{0.11}         & \tt{9.25}              & \tt{2:35:06}            \\
    Scaled masked cKDTree & \tt{0.03}         & \texttt{\textbf{0.70}} & \texttt{\textbf{11:35}} \\ \midrule
    TNN GPU               & \tt{7.00}         & \texttt{\textbf{0.16}} & \texttt{\textbf{2:43}}  \\

    Linear search GPU     & \tt{- }           & \tt{2.55}              & \tt{42:28}              \\
    \bottomrule
  \end{tabular}
\end{table*}
\subsection{Comparison with KDTrees} \label{sec:kdtree}

We compared our method to a KDTree implementation that extends the one proposed by Scikit Learn \cite{scikit-learn}. The goal here is to compare the two methods on concrete examples and see which one is better suited for finding nearest neighbors in a Machine Learning context.
We evaluated a version in pure python (\textit{Scaled Masked KDTree}) and an optimized one in Cython (\textit{Scaled Masked cKDTree}) [Tab. \ref{tab:comparison}].
We changed the metric in KDTrees to have a scalable query and retrieve only the point after a given time window.
One can see that, on the original data set, the scaled and masked KDTree implementation in pure python is on par with the linear search, which highlights that KDTrees are not designed to work with masked data.
This was not their intended use, as KDTrees explore in priority cells close to the query point.
However, if one dimension is masked, many of these cells contain no valid candidates resulting in a substantial decrease in performance as more cells need to be explored [Fig. \ref{fig:kdtree_masked}].
An extensive ablation study is made in the supplementary material showcasing this phenomenon.
This bad performance can be mitigated by using an optimized implementation in Cython, and we see that on CPU, the cKDTree is the fastest method. However, our approach still offers a substantial increase even when used on CPU where it cannot benefit as much from the GPU parallelization capabilities. Furthermore, the KDTree algorithm is not appropriate for running on GPU as it requires many non-parallelizable operations.
The real advantage of our method is that it runs efficiently on GPU, which allows it to take advantage of the parallelization speedup. Comparing the optimized version of cKDtree on CPU to TNN on GPU, one can see that TNN seems to offer the best running time. Furthermore, a GPU implementation is particularly well suited for Machine Learning applications where most of the data is processed on the GPU, so it should benefit additionally from the spared communication time.

\subsection{Conclusion}
High altitude wind nowcasting differs from weather forecasting. In the first few hours, extrapolation of high-quality measurements is still the most efficient approach because of the persistence of weather phenomena and because weather forecasts use numerical grids whose resolution is too large.
Working on unstructured data measured along the trajectories of airplanes offers another challenge, as creating contexts for a prediction is not an easy task:
Restricting the context to a small set of neighbors is mandatory to reduce the different models' costs in time and space. This alone reduces by almost two orders of magnitude the duration of the models' training epochs. Nevertheless, finding a good context is not straightforward as data recorded in the future has to be masked while depending on the scaling of the different dimensions. Moreover, traditional methods are not well suited to work with masked data and encompass many non-tensorial operations, making it difficult to adapt to GPU. On the contrary, TNN splits the dataset in a natural way and runs efficiently on GPU.

We proposed a novel algorithm for searching the $k$-Nearest Neighbors
($k$-NN) for the specific case of points organized along piece-wise
linear trajectories in a Euclidean space, which allows masking
points along a given dimension. The general required property is that
the data sets admit a coverage with cylinders. This algorithm is
formulated with parallelizable tensorial operations and works well on GPU. A Pytorch implementation is provided \cite{Paszke2019}, making its
integration easy for most Machine Learning pipelines. It allowed us to reach a substantial increase in efficiency in the case of this study.
By using this algorithm as a stepping stone, additional trials of the method using more complex models will be performed to increase the precision of the extrapolation scheme while remaining efficient.

\paragraph{Acknowledgement}
Arnaud Pannatier was supported by the Swiss Innovation Agency
Innosuisse under grant number 32432.1 IP-ICT -- ``MALAT: Machine
Learning for Air Traffic.''

\bibliographystyle{siam}
\bibliography{main}
\end{document}